\documentclass[runningheads]{llncs}
\usepackage[T1]{fontenc}
\usepackage{graphicx}
\usepackage{subcaption}
\usepackage{amsmath}
\usepackage{amssymb}
\usepackage{bm}
\usepackage{multirow}
\usepackage{array}
\usepackage[hidelinks]{hyperref}
\usepackage{cleveref}
\usepackage[table]{xcolor}

\newcommand{\added}[1]{#1}
\newcommand{\removed}[1]{}
\begin{document}
\title{Selecting haptic guidance models in teleoperation: guidelines from a comparative user study}
\titlerunning{Selecting haptic guidance models in teleoperation}
\author{Alexis Boulay\inst{1,2}\orcidID{0009-0001-1125-6971} \and
Margot Vulliez\inst{2}\orcidID{0000-0002-2476-8327} \and
David Daney\inst{2}\orcidID{0000-0001-8538-5875}}
\authorrunning{A. Boulay et al.}
\institute{Farm3, Besançon, France \and
Auctus Team, Inria, Talence, France\\
\email{name.lastname@inria.fr}}

% \author{First Author\inst{1,2}\orcidID{0000-1111-2222-3333} \and
% Second Author\inst{2}\orcidID{1111-2222-3333-4444} \and
% Third Author\inst{2}\orcidID{2222--3333-4444-5555}}

% \institute{Institution A, City A, Country A \and
% Institution B, City B, Country B\\
% \email{dummy@mail.com}}

%
\maketitle              % typeset the header of the contribution
\begin{abstract}
Haptic guidance in teleoperation enhances operator performance through force feedback. This paper presents guidelines to select the most appropriate model considering the task, the environment and the operator. We define a unified \added{formulation} expressing most common models (spring-damper, potential field, and guiding tube) as variations of a stiffness-damping system with model-specific guiding functions. We conducted a user study comparing the three classical models across six scenarios with varying environmental conditions in a vertical farming task. Results show no universally superior model: spring-damper excels in cluttered environments, potential field in free spaces (but it shows risks near obstacles), and guiding tube offers a balanced compromise. We propose novel objective metrics to evaluate the interaction, and show that guiding force magnitude correlates with comfort and trust scores. These findings provide practical model selection guidelines through environmental characteristics and real-time evaluation metrics.

\end{abstract}
\section{Introduction}

% Context
Robotic teleoperation allows a human operator to achieve a task in a remote environment through the control of a robotic system. While this technique provides a safe way to perform tasks in hazardous or inaccessible environments, it can lead to reduced performance and increased cognitive workload due to limited or imperfect remote sensory feedback. In this study we focus on the example of vertical farming, where the varying sizes and positions of plants create dynamic and cluttered environments, making teleoperation more challenging and highlighting the need for effective assistance, such as haptic guidance.

Introduced as virtual fixtures~\cite{rosenberg1993virtual}, also known as active constraints~\cite{davies2006active}, haptic guidance helps human operators by transmitting forces that guide their movements towards desired positions or away from forbidden areas. Haptic guidance is composed of two elements: the virtual object, which represents the target of the guidance and can be either an attractive goal/path or a repulsive obstacle, and the guidance model, which determines the force profile based on the distance to the chosen virtual object. While virtual objects can be explicitly defined from knowledge of the task and the environment, the choice of the guidance model is not straightforward and could further depend on user preferences.

% Literature gap

Recent literature review on haptic guidance~\cite{bowyer2013active} presents different types of guidance, but no guidelines for choosing the best model exists for a given situation. The state of the art also lacks a generic formulation that would enable the comparison of guidance models across different use cases.

% Methodology
To address these gaps, we first define a unified formulation for haptic guidance that encompasses the three main guidance models found in the literature. We then propose a set of objective metrics to evaluate the quality of human-guidance interaction based on four criteria: performance, comfort, safety, and trust. \removed{Since the human operator is a key element in the guidance loop,} We conduct a user study to compare how operators interact with the different guidance models under various environmental conditions and to evaluate the proposed metrics.

\begin{figure}[!ht]
    \centering 
    \includegraphics[width=0.9\linewidth]{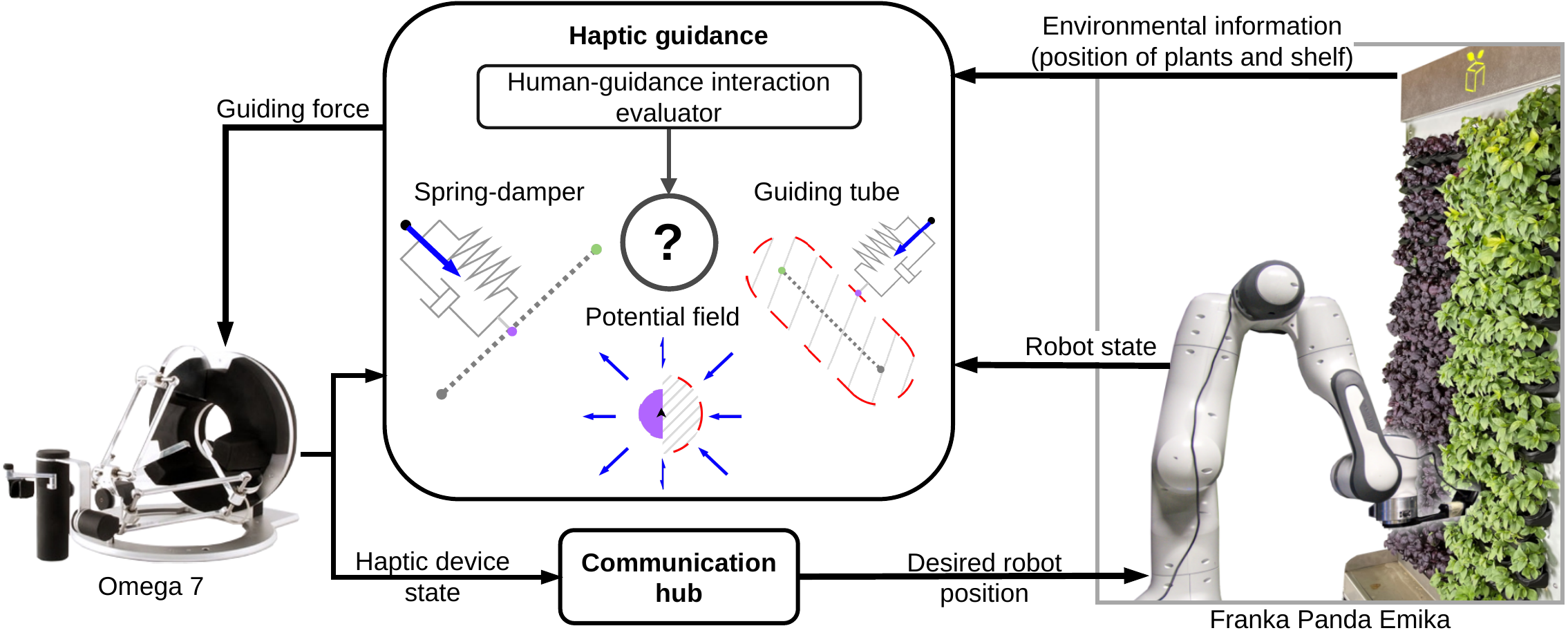}
    \caption{Overview of the proposed approach: an interaction evaluator is embedded within the haptic guidance module to online select the most appropriate guidance model based on task, environment, and interaction quality evaluation.}
    \label{fig:farm_shelf_dgrm}
\end{figure}

% Contributions
\removed{This paper presents a unified formalism expressing main haptic guidance models as variations of a stiffness-damping system with model-specific guiding functions.
We propose a set of objective metrics for real-time evaluation of human-guidance interaction quality and establish practical guidelines for model selection through a comparative user study. As illustrated in \cref{fig:farm_shelf_dgrm}, these contributions enable the selection of the most appropriate haptic guidance model based on task requirements, environmental characteristics, and real-time interaction quality evaluation.}

\added{This paper proposes guidelines for selecting the most appropriate haptic guidance model based on task requirements, environmental characteristics, and real-time evaluation of interaction quality, as illustrated in \cref{fig:farm_shelf_dgrm}. This contribution is built upon three key developments: 
\begin{itemize}
    \item A unified formulation that encapsulates classical haptic guidance models as a stiffness-damping system with a model-specific guiding function, enabling direct comparison across main existing models of the literature.
    \item The development of a set of objective metrics for real-time evaluation of human-guidance interaction quality from force-motion observations, which are experimentally validated against to user experience through standard questionnaires.
    \item A user study designed to compare guidance models under varying task conditions.
\end{itemize}
}

% Plan
\Cref{sec:models} presents the unified \added{formulation} and describes the three main guidance models along with the proposed interaction evaluation metrics. \Cref{sec:study} details the user study methodology. \Cref{sec:results} presents and discusses the results, comparing guidance models across scenarios and validating the proposed metrics. Finally, \cref{sec:conclusion} concludes and outlines future work.

\section{Haptic guidance models}
\label{sec:models}
\subsection{A unified definition of force guidance}

\begin{figure}[!ht]
    \centering 
    \includegraphics[width=0.8\linewidth]{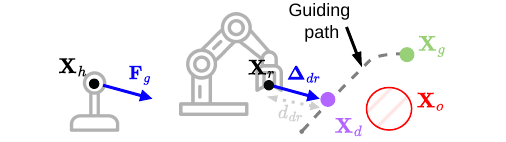}
    \caption{A guiding force $\bm{F}_g$ is generated on the haptic interface based on the guidance model and the distance between the robot position $\bm{X}_r$ and a virtual object (goal $\bm{X}_g$, path point $\bm{X}_d$, or obstacle $\bm{X}_o$).}
    \label{fig:intro_guide}
\end{figure}

Haptic guidance, as depicted in \cref{fig:intro_guide}, is a force feedback, either attractive or repulsive, provided to help the operator perform a task.

In the literature, various types of haptic guidance have been proposed, such as guidance that assists users in following a trajectory~\cite{raiola2018_co-manipulation}, guidance that helps users avoid obstacles~\cite{binet2018_using}, or guidance that prevents operators from exceeding defined limits~\cite{li2005_constrained}. The expression of haptic guidance forces varies considerably across publications, but we propose a unified \added{formulation} for these approaches.

In the following equations, the force feedback $\bm{F}_g\in\bbbr^3$ is written considering Cartesian positions only, but it can be extended to $\bbbr^6$ by computing torque considering orientation distance such as geodesic distance on a Lie group~\cite{alberto2023predictive}.

Given the position of the teleoperated robot $\bm{X}_r\in\bbbr^3$ and the position of a virtual object $\bm{X}_d\in\bbbr^3$, the guiding force depends on the distance $d_{dr} \in\bbbr$ and the unit direction vector $\bm{\Delta_{dr}}\in\bbbr^3$:

\begin{equation}
\bm{\Delta_{dr}} = \frac{\bm{X}_r - \bm{X}_d}{d_{dr}} \text{ with } d_{dr} = ||\bm{X}_r - \bm{X}_d||_2
\end{equation}

Any guiding force can be unified through a stiffness-damping expression, with the stiffness and damping gains $\bm{K}_g\in\bbbr^{3\times 3}$ and $\bm{B}_g\in\bbbr^{3\times 3}$, and the velocity of the haptic device $\dot{\bm{X}}_h\in\bbbr^3$. The behavior of the guiding force with respect to the distance $d_{dr}$ will vary according to the chosen guidance model through a guiding function $f_{gm}(d_{dr})\in\bbbr$:
\begin{equation}
    \bm{F}_g = f_{gm}(d_{dr}) \bm{K}_g \bm{\Delta_{dr}} - \bm{B}_g \dot{\bm{X}}_h
    \label{eq:fg}
\end{equation}

All the terms in \cref{eq:fg} are expressed in the robot world frame. The final force feedback is transformed back to the haptic device frame before being applied to the operator. The gains $\bm{K}_g$ and $\bm{B}_g$ can be defined to privilege certain directions of guidance if needed, but in the following we consider them as diagonal matrices of constant values $k_g\in\bbbr$ and $b_g\in\bbbr$ respectively.

\added{This representation facilitates the comparison of different guidance models through their guiding function $f_{gm}(d_{dr})$, and it provides a common framework for tuning the parameters of the guidance models.}

\subsection{Existing guidance models}

\begin{figure}[!ht]
    \centering
    \begin{subfigure}[b]{0.32\textwidth}
        \centering
        \includegraphics[width=\linewidth]{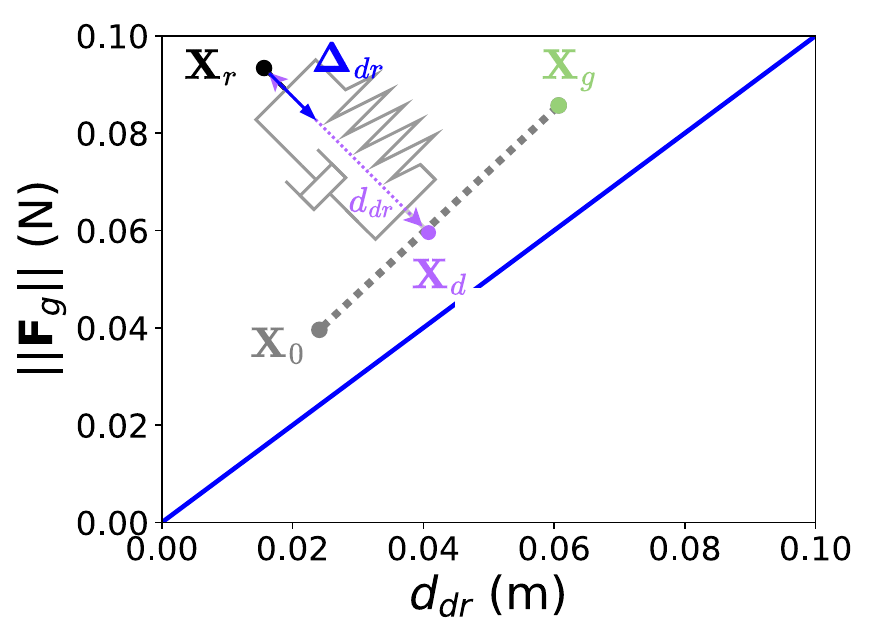}
        \caption{Spring-damper}
    \end{subfigure}
    \hfill
    \begin{subfigure}[b]{0.32\textwidth}
        \centering
        \includegraphics[width=\linewidth]{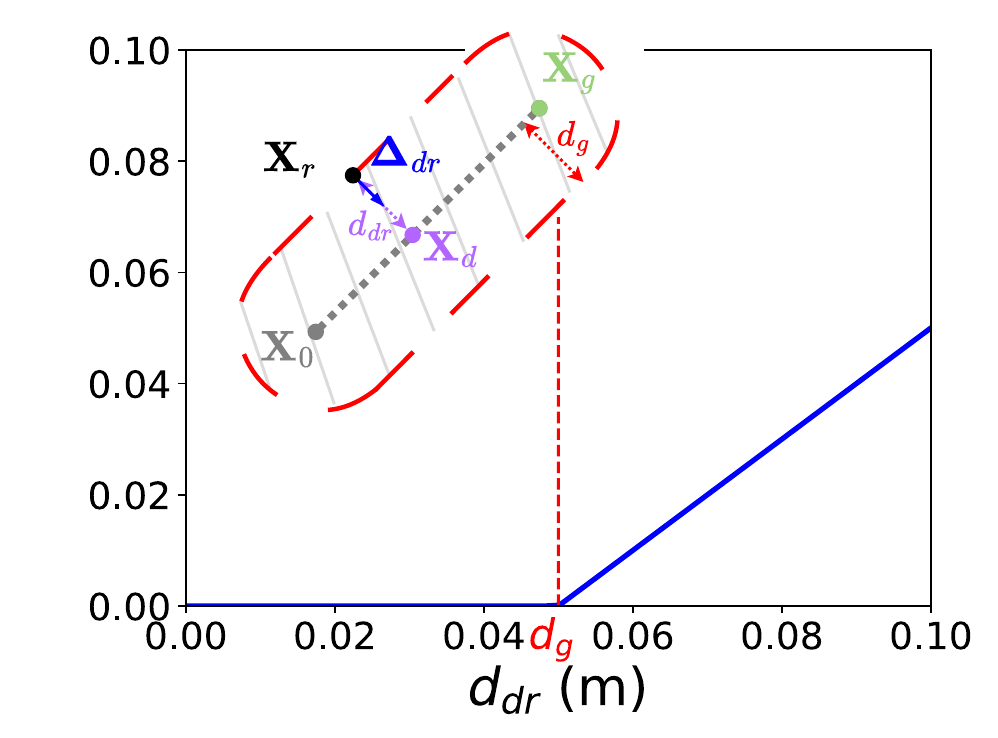}
        \caption{Guiding tube}
    \end{subfigure}
    \hfill
    \begin{subfigure}[b]{0.302\textwidth}
        \centering
        \includegraphics[width=\linewidth]{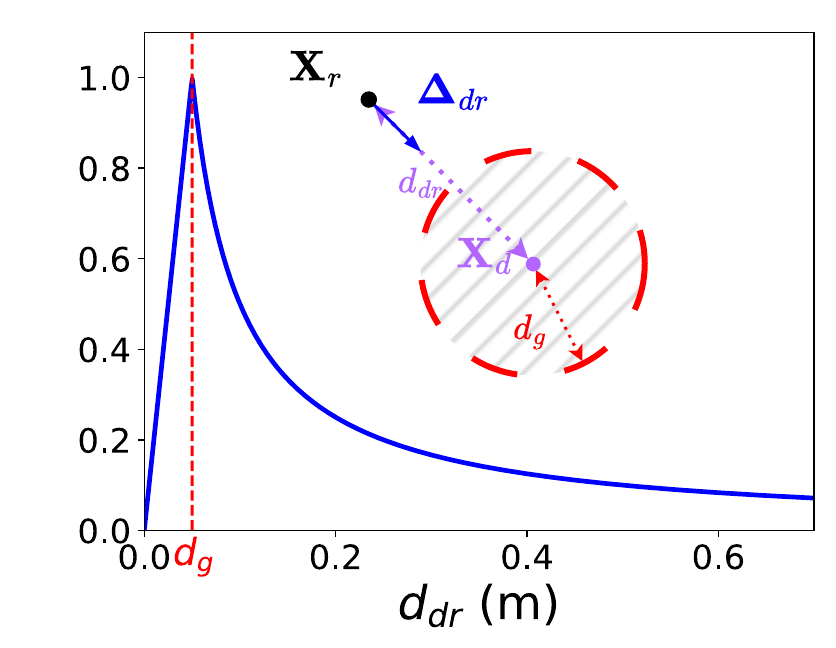}
        \caption{Potential field}
    \end{subfigure}
    \caption{\small Force profiles for the different guidance models: (a) force is proportional to distance; (b) force is equal to zero until radius $d_g$, then it is proportional; (c) force increases near attraction point, decreases below $d_g$. The forces profiles are represented in static conditions (no damping effect) with $k_g=1$.}
    \label{fig:guide_models}
\end{figure}

In the literature, the function $f_{gm}(d_{dr})$ is not explicitly defined in most papers, and only the overall behavior of the guiding force is described. However, we identified three main guidance models, depicted in~\cref{fig:guide_models}: the spring-damper model, which prevents path deviation; the potential field model, for multiple attractive/repulsive objects; and the guiding tube model, which prevents area exit. For each identified model, we define a corresponding guiding function $f_{gm}(d_{dr})$.

The \textbf{spring-damper} model is used for precisely guiding movements in tasks like surgical teleoperation~\cite{shi2023_haptic} or path-following~\cite{restrepo2020_toward,zlajpah2019_virtual}. Usually combined with a path as virtual object, the force increases proportionally with distance:
\begin{equation}
    f_{gm}(d_{dr}) = f_{sd}(d_{dr}) = d_{dr}
\end{equation}

The \textbf{potential field} model suits tasks with multiple goals or obstacles, such as grasping~\cite{abifarraj2020_haptic} or navigation with obstacle avoidance~\cite{binet2018_using}. The force increases when approaching the point, then decreases below a threshold $d_g$ (when the target is reached). Attractive/repulsive forces are set through the stiffness sign:
\begin{equation}
    f_{gm}(d_{dr}, d_g) = f_{pf}(d_{dr}, d_g) = \frac{d_{dr}}{d_g}e^{1-\frac{d_{dr}}{d_g}}
\end{equation}

The \textbf{guiding tube} model ensures operators stay within areas, maintaining robot reachable space~\cite{zolotas2021_polytopes} or safe paths~\cite{muhlbauer2022_multi}. When used with a path, the force is zero under a radius $d_g$, then proportional to the distance to the tube limit:
\begin{equation}
    f_{gm}(d_{dr}, d_g) = f_{gt}(d_{dr}, d_g) = \begin{cases}
    0 & \text{if } d_{dr} \leq d_g \\
    d_{dr} - d_g & \text{if } d_{dr} > d_g
    \end{cases}
\end{equation}

\added{While the spring-damper model and the potential field model guide users towards a specific motion, the guiding tube model imposes a restriction to stay within a given workspace. However, when used pairwise with a path, users can stay in contact with the guiding tube to follow the path, making it a good model for guidance too.}

In addition to the guiding function, the parameter tuning varies depending on the model. Spring-damper stiffness must be low enough that the users can overcome path errors. Guiding tube needs high stiffness to make the user stay within the tube of radius $d_g$. Potential field requires per-object tuning of stiffness and $d_g$ considering object size and importance for the task.

While typical use cases can be identified for each model, no generic selection guidelines are available. To the best of our knowledge, there is no study certifying that one model is more suited to a particular situation. The literature also lacks criteria and metrics to assess the efficiency of a proposed haptic guidance regarding the operator. 

\subsection{Evaluating human-guidance interaction}
\label{subsec:hgi}

The quality of the interaction between the human operator and the haptic guidance assistance is crucial for the success of the teleoperation task and can serve as a good indicator of the guidance model's suitability. A well-designed haptic guidance should enhance the operator's performance~\cite{olsen2003metrics} while ensuring their comfort~\cite{prewett2010managing}, safety~\cite{zanchettin2016_safety}, and trust in the system~\cite{benhabib2021_quantify,nikolaidis2017human}. While subjective metrics such as questionnaires provide valuable insights, objective metrics measured during task execution can offer complementary real-time assessment of interaction quality.

We propose four novel metrics ($\bm{M_{P_2}}$, $\bm{M_{C}}$, $\bm{M_{T_1}}$, and $\bm{M_{T_2}}$) to evaluate human-guidance interaction alongside two commonly used metrics ($\bm{M_{P_1}}$ and $\bm{M_{S}}$). These metrics are based on our four criteria: performance, comfort, safety, and trust.

\begin{figure}[!ht]
    \centering
    \begin{subfigure}[b]{0.49\textwidth}
        \centering
        \includegraphics[height=2cm]{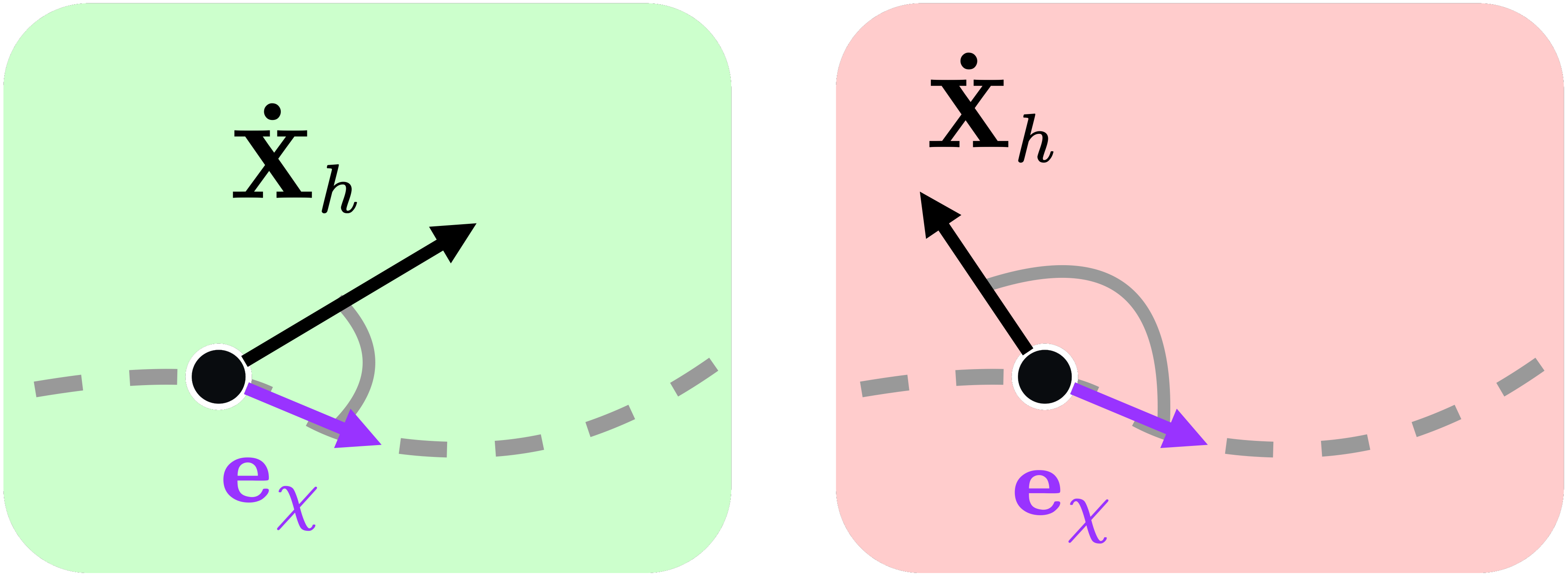}
        \caption{\added{Path-tracking accuracy ($\bm{M_{P_2}}$): following the optimal path yields good performance (left), while deviating from it yields poor performance (right).}}
    \end{subfigure}
    \hfill
    \begin{subfigure}[b]{0.49\textwidth}
        \centering
        \includegraphics[height=2cm]{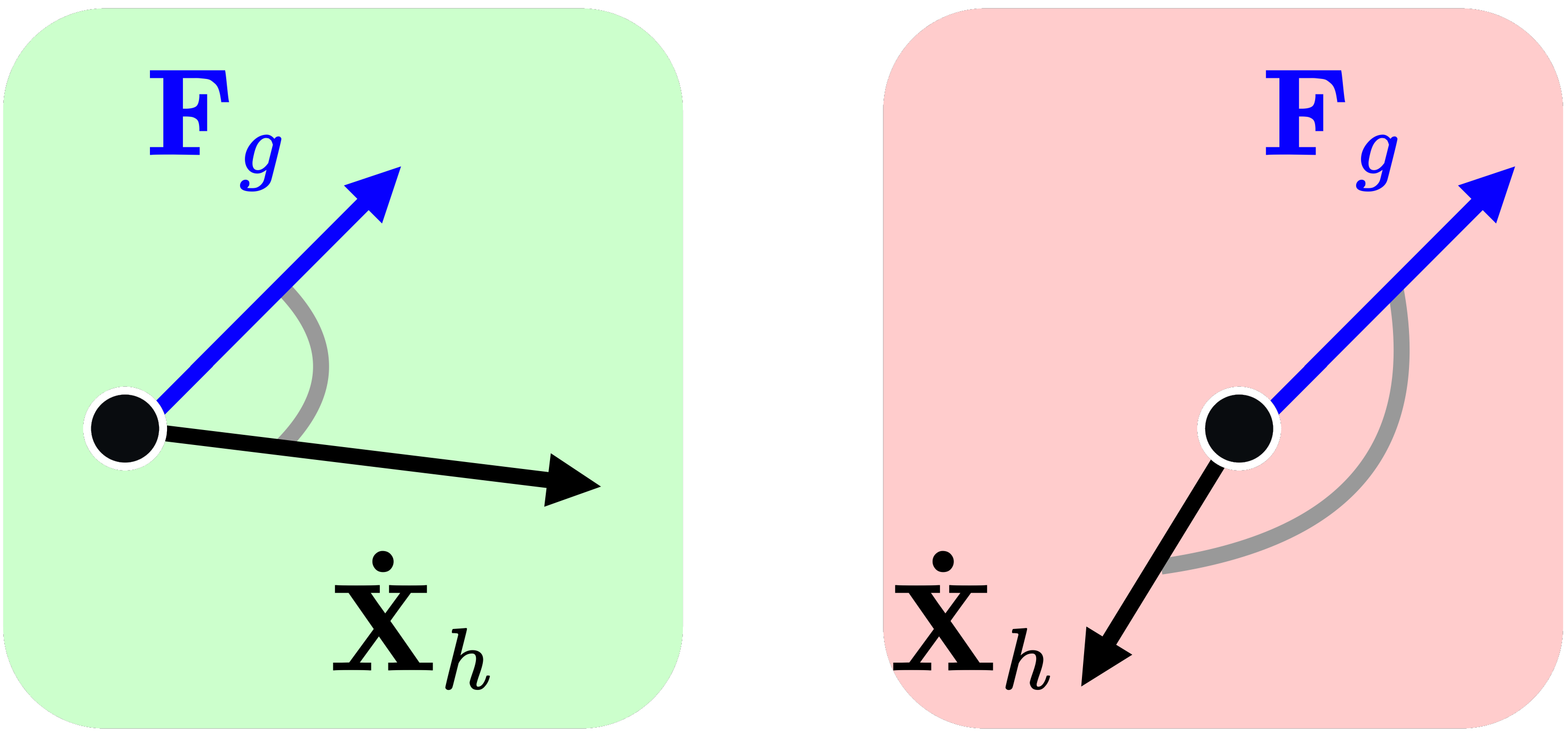}
        \caption{\added{Force-tracking accuracy ($\bm{M_{T_2}}$): moving with the guidance force indicates trust (left), while moving against it indicates distrust (right).}}
    \end{subfigure}
    \caption{\added{Illustration of path-tracking accuracy for spring-damper/guiding tube and force-tracking accuracy for potential field.}}
    \label{fig:metrics_illustration}
\end{figure}

\textbf{Performance} is often assessed by recording the task completion time $\bm{M_{P_1}$}. For spring-damper and guiding tube guidance models, \added{we hypothesize} that performance \added{could} be related to path-tracking accuracy $\bm{M_{P_2}$}, illustrated in \cref{fig:metrics_illustration}; if guidance is effective, the guide is optimal for the task and the user should follow it closely. We describe the guiding path $\bm{\chi}$ as a discrete set of waypoints $[\bm{X}_\chi(1), \dots, \bm{X}_\chi(N)]$, where $\forall i\in[\![1;N-1]\!]$, $\bm{X}_\chi(i)\in\bbbr^3$ is a point coordinates and $\bm{e}_\chi(i) = \bm{X}_\chi(i+1)-\bm{X}_\chi(i)\in\bbbr^3$ approximates the tangent at this point.
At time $t$, we consider the velocity of the haptic interface projected in the robot world frame $\bm{\dot{X}}_{h}(t)$, the position of the robot $\bm{X}_r(t)$, and we define the orthogonal projection of this position onto the guiding path as the closest path point with index $k\in[\![1;N]\!]$, $\bm{X}_\chi(k) = \bm{proj^\perp _{\chi}}(\bm{X}_{r}(t))$. Then $\bm{M_{P_2}}$ can be measured in real-time and is written as:
\begin{equation}
  \bm{M_{P_2}}(t)= \frac{\bm{e}_{\chi}(k)^T\bm{u}_{h}(t)}{||\bm{e}_\chi(k)||_2} \text{ with } \bm{u}_{h}(t)=\frac{\bm{\dot{X}}_{h}(t)}{||\bm{\dot{X}}_{h}(t)||_2}
\end{equation}

\textbf{Comfort} \added{is hypothesized} to be assessed by measuring the guiding force magnitude $\bm{M_{C}$}; the higher the force, the more the user has to resist it and the less comfortable the guidance is:
\begin{equation}
    \bm{M_{C}}(t) = ||\bm{F}_g(t)||_2
\end{equation}

\textbf{Safety} is often assessed by recording the minimum distance to the nearest obstacle from an obstacle set $\{\bm{X}_o\}$~\cite{binet2018_using}; the larger this distance, the safer the interaction is.
\begin{equation}
    \bm{M_{S}}(t) = \min_{\bm{X}_{oi}\in\{\bm{X}_o\}} ||\bm{X}_r(t) - \bm{X}_{oi}||_2
\end{equation}

\textbf{Trust} in the haptic guidance system \added{could be assessed trough force-motion agreement metrics}, that are evaluated differently depending on the guidance model. For spring-damper and guiding tube models, we measure the guiding force magnitude $\bm{M_{T_1}$}; \added{we hypothesize that} a higher force indicates that the user is moving away from the guiding path and do not trust it.
\begin{equation}
    \bm{M_{T_1}}(t) = ||\bm{F}_g(t)||_2
\end{equation}

For the potential field model, \added{agreement can be measured by} the force-tracking accuracy $\bm{M_{T_2}$}, illustrated in \cref{fig:metrics_illustration}; if the user doesn't trust the guidance, \added{we hypothesize that} they will tend to go on a different direction than the one indicated by the guidance force:
\begin{equation}
    \bm{M_{T_2}}(t) = \frac{\bm{F}_g(t)^T \bm{u}_h(t)}{||\bm{F}_g(t)||_2} \text{ with } \bm{u}_h(t)=\frac{\bm{\dot{X}}_h(t)}{||\bm{\dot{X}}_h(t)||_2}
  \label{eq:t2}
\end{equation}

As these new metrics are not commonly used in the literature, we propose exploiting the results of the user study, conducted to compare the different guidance models, to also evaluate whether these metrics effectively assess human-guidance interaction quality by comparing them with subjective evaluations.

\section{A comparative user study of guidance models}
\label{sec:study}
\subsection{Vertical farming as use case scenario}
In order to evaluate the different haptic guidance models (spring-damper, potential field, and guiding tube) under different environmental configurations and to validate the definition of metrics about human-guidance interaction, we designed a user study based on a real-world scenario: taking photos of plants to monitor their growth in teleoperated vertical farming.

In vertical farming, regular monitoring of plant growth is essential. Teleoperation prevents disturbances in the climate-controlled growing environment/room caused by human comings and going. Preserving the human expertise within the teleoperation loop also allows performing a task which can hardly be automated because of the environment's dynamic and cluttered nature. 

\begin{figure}[!ht]
    \centering
    \begin{subfigure}[b]{0.41\textwidth}
        \centering
            \includegraphics[width=\textwidth]{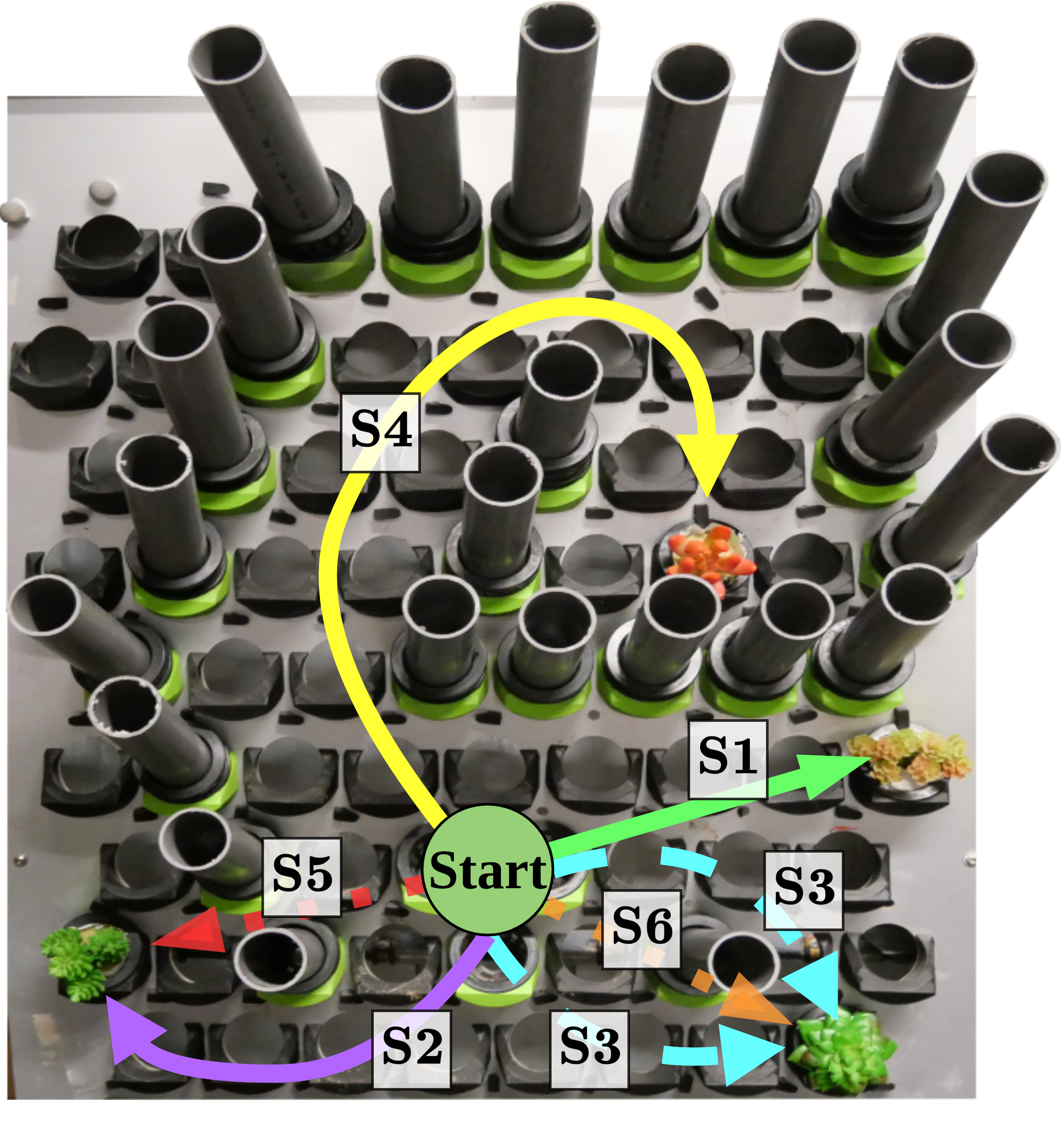}
            \caption{}
    \end{subfigure}
    \hfill
    \begin{subfigure}[b]{0.58\textwidth}
        \centering
        \includegraphics[width=\textwidth]{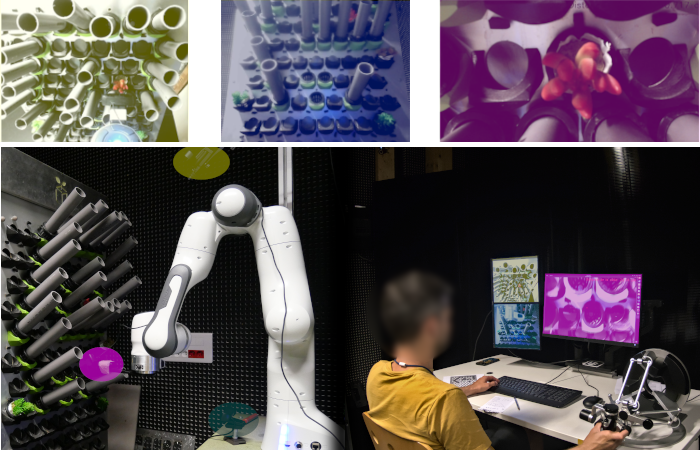}
        \caption{}
    \end{subfigure}
    \caption{Experimental setup and scenarios for the user study. (a) The 6 scenarios used: $\bm{S1}$ is a straight line, $\bm{S2}$ requires obstacle avoidance, $\bm{S3}$ allows avoidance from two sides, $\bm{S4}$ is highly constrained, and $\bm{S5}$/$\bm{S6}$ use guides that ignore obstacles. (b) Teleoperation setup with a haptic interface and visual feedback from 3 cameras, used by the operator to remotely control the robot and take photos of plants on a farming shelf.}
    \label{fig:env_scenarios}
\end{figure}

We replicate such a complex interaction environment for the user study, where we ask operators to reach a series of plants arranged on a farming shelf and to take a photo of each plant using a camera mounted on the end-effector of the teleoperated robot (with 2 cm positioning precision). To achieve the task, the operator's visual feedback is provided by three cameras: the main one mounted on the robot end-effector, one with a top view of the shelf, and one with a bottom view. On the shelf, we place several PVC tubes as obstacles to simulate the cluttered environment. We define six different scenarios $\{\bm{S1},\ \dots,\ \bm{S6}\}$ illustrated in \cref{fig:env_scenarios}. $\bm{S1}$, $\bm{S2}$, and $\bm{S4}$ present different levels of environment clutter (respectively free, with one obstacle, very constrained); $\bm{S3}$ illustrates ambiguous assistance where the user must choose between two sides to avoid the obstacle when the path guide is randomly pre-generated on one side; and $\bm{S5}$ and $\bm{S6}$ represent use cases where the virtual object is defined without considering the obstacles (poorly defined assistance).

All participants perform the task for the six scenarios and under four different guidances modalities: without haptic guidance, with spring-damper or guiding tube model using a path as the virtual object, and with potential field model using goal and obstacle points as virtual objects. \added{Each guidance model is, therefore, tested 6 times per participants.} The participants are given a random scenario and then have to perform the task successively once for each condition in a random order without prior knowledge of it.

\subsection{Evaluation protocol}
To evaluate the interaction between the human operator and the haptic guidance, we combine the objective metrics ($\bm{M_{P_1}$}, $\bm{M_{P_2}$}, $\bm{M_{C}$}, $\bm{M_{S}$}, $\bm{M_{T_1}$}, $\bm{M_{T_2}$}) defined in \cref{subsec:hgi} with subjective assessments. This dual approach enables us to validate whether the proposed objective metrics effectively capture the operator's experience.

For each task attempt (scenario and condition), participants complete two questionnaires. The NASA-RTLX~\cite{georgsson2019nasa} evaluates cognitive workload across six dimensions (mental demand, physical demand, temporal demand, performance, effort, and frustration), providing insights into the comfort and perceived performance of the interaction. The Muir questionnaire~\cite{muir1996_trust} assesses trust in the haptic guidance system through questions about its reliability, predictability, and dependability.

When participants complete a scenario under all four conditions (no guidance, spring-damper, guiding tube, and potential field), they rank their preferred conditions. This ranking provides a holistic assessment of user satisfaction that combines all interaction quality criteria.

During each task attempt, the objective metrics are recorded continuously. The mean values of time-dependent metrics are computed post-task for statistical analysis. By comparing these objective measurements with subjective questionnaire responses, we can determine whether the proposed metrics effectively reflect the operator's perceived interaction quality and identify which objective metrics correlate most strongly with subjective assessments.

\subsection{Teleoperation bilateral controller}
\begin{figure}[!ht]
  \centering
    \includegraphics[width=\textwidth]{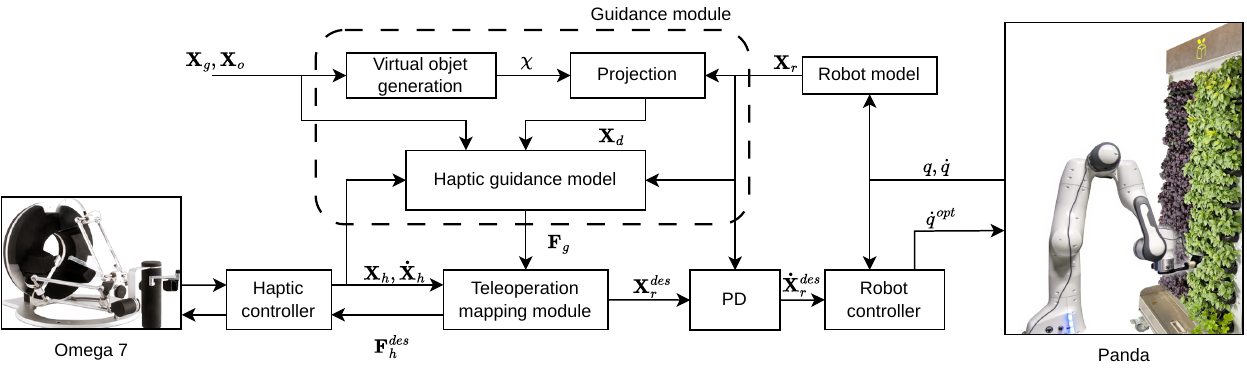}
    \caption{Bilateral teleoperation control diagram.}\label{fig:pgm_control_diag}
\end{figure}

The teleoperation framework uses ROS for communication \added{between two local robot and haptic controllers running at 1 kHz.}
The haptic interface transmits position/velocity ($\bm{X}_h,\bm{\dot{X}}_h$) and receives force ($\bm{F}_{h,des}$). \added{Exchanging force variables through a communication with non-deterministic time delays as ROS could induce instability in the teleoperation feedback loop. Since there is no direct force sensing but only haptic guidance being transmitted as feedback, the haptic force computation can be done locally within the haptic control loop. Acting as a stiffness-damping proxy, the force computation only depends on exchanged robot position information, which cannot jeopardize stability of the teleoperation framework~\cite{jorda2022_local}.} A mapping module applies spatial transformations between the haptic and robot frames (scale $s_{h\to r}=5.0$, matrix $\bm{T}_{h\to r}$). Desired robot position feeds a PD controller generating joint velocity commands via constrained quadratic optimization~\cite{joseph2020}. A guidance module (\cref{fig:pgm_control_diag}) switches between models, computing $\bm{F}_g$ using robot and haptic states as well as environment information (position of virtual objects).

Parameters of each guidance model are tuned empirically to provide effective assistance without hindering task completion. \added{Parameter values for each guidance model are selected from expert feedback during preliminary tests to ensure effective assistance while avoiding excessive forces or hazardous motions (e.g., oscillations). For the potential field model, force-field map visualizations are used to ensure that users could navigate the environment easily (with a force below $5$ N) without being trapped in local minima. For the guiding tube model, the radius $d_g$ is set so that obstacles are excluded from the tube in every scenario, and the gain are set to create a perceptible discontinuity in user force perception at the tube boundary. For the spring-damper model, the gains are set so that users would experience the maximum force that the haptic interface can provide ($12$ N) from a deviation of $2$ cm from the path. For all models, damping gains are then set to limit oscillations when quick movements are made around the path/target.}
For the spring-damper model, $k_g=600$ N/m and $b_g=17$ Ns/m; for the guiding tube model, $k_g=2000$ N/m, $b_g=7$ Ns/m, and $d_g=0.08$ m; for the potential field model, the goal used $k_g=20$ N/m and $d_g=0.08$ m, while obstacles used $k_o=25$ N/m, $d_o=0.02$ m, and $b_g=3$ Ns/m.

\added{Although per-participant tuning could improve absolute performance regardless of guidance model, it would make model-to-model comparisons less relevant, since the behavior of the guidance would be different for each participant.}

% \begin{table}[!ht]
% \centering
% \caption{Guidance parameters.}\label{tab:guidance_params}
% \begin{tabular}{|l|c|c|c|}
% \hline
% Model & Spring-damper & Guiding tube & Potential field \\
% \hline
% $k_g$ (N/m) & 600 & 2000 & goal: 20, obst: 25 \\
% $b_g$ (Ns/m) & 17 & 7 & 3\\
% $d_g$ (m) & N/A & 0.08 & goal: 0.08, obst: 0.02 \\
% \hline
% \end{tabular}
% \end{table}

\section{Results and discussion}
\label{sec:results}
\removed{Statistical analysis (R 4.4.1) used ANOVAs with Greenhouse-Geisser correction, $p<0.05$ significance.}
\added{Statistical analyses are conducted using RStudio (R 4.4.1). Normality and homogeneity of variance are assessed using the Shapiro-Wilk and Levene tests, respectively. ANOVAs are performed, with Greenhouse-Geisser correction applied when Mauchly's test indicated a violation of sphericity. Statistical significance was set at $p<0.05$. When significant effects are observed, Bonferroni-corrected post-hoc pairwise comparisons are performed to identify which conditions differed significantly.}

Twenty-eight volunteers participated; eight were excluded because data-recording issues \added{produced incomplete datasets in some random trials - packet loss in the recording communication} (final sample: 8F, 12M). The study was approved by the ethical committee, and all participants signed a consent form.

\begin{figure}
    \centering
    \includegraphics[width=0.9\textwidth]{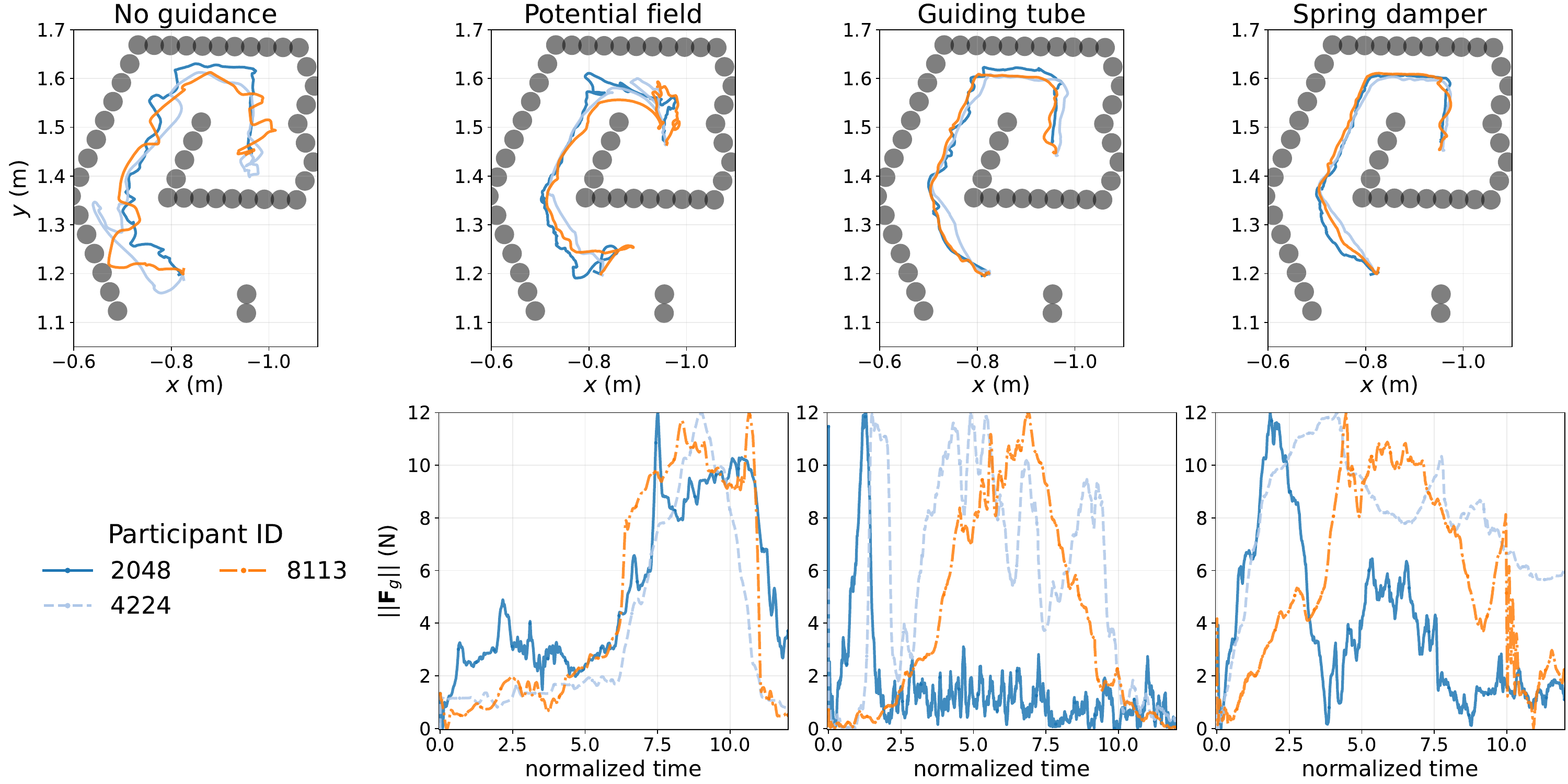}
    \caption{\added{Trajectory of three participants in scenario $\bm{S4}$ under the different guidance models. The resulting guiding forces are plotted on the bottom for each model in normalized time.}}\label{fig:ex_traj_s4}
\end{figure}

Participants photographed plants while avoiding obstacles after completing three training scenarios in the same environment as the main scenarios. \added{An example of the resulting trajectories for three participants in scenario $\bm{S4}$ under the different guidance models is illustrated in \cref{fig:ex_traj_s4}. } 

\subsection{Guidance model comparison}

\begin{figure}[!ht]
    \centering
    \begin{subfigure}[t]{0.49\textwidth}
        \includegraphics[width=\textwidth]{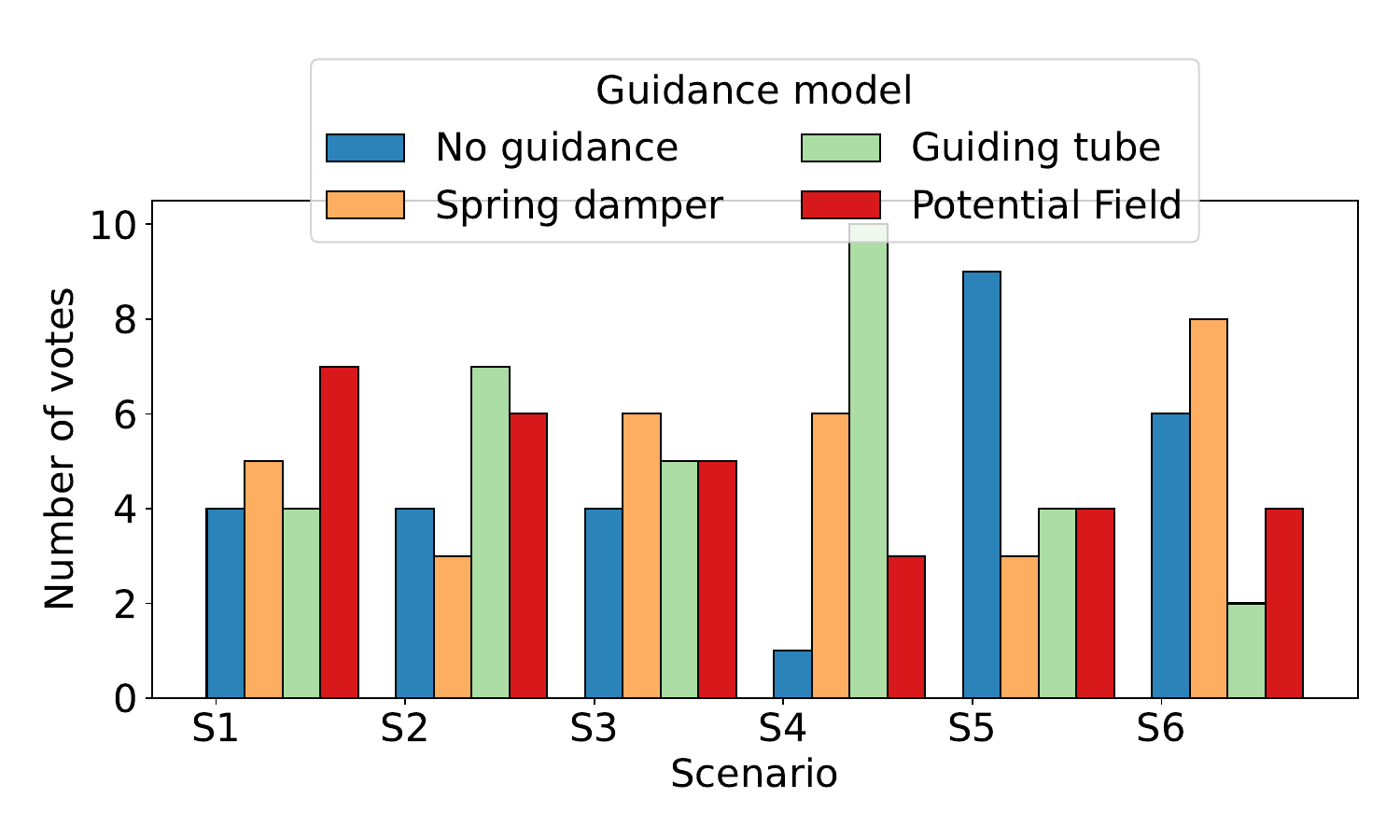}
        \caption{Frequency of first-place}
    \end{subfigure}
    \begin{subfigure}[t]{0.39\textwidth}
        \includegraphics[width=\textwidth]{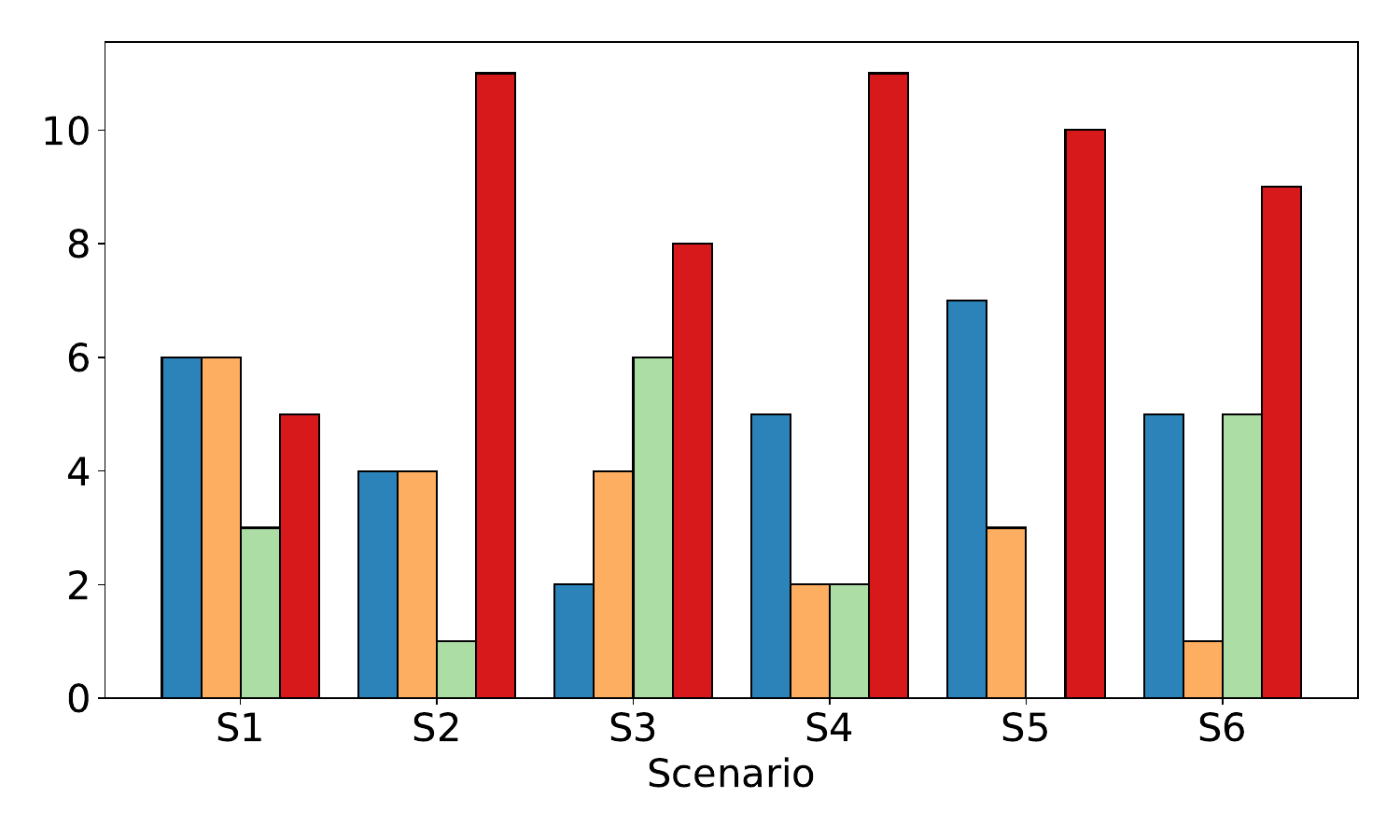}
        \caption{Frequency of last-place}
    \end{subfigure}
    \caption{Participants' ranking of guidance models across different scenarios.}\label{fig:pgm_classification}
\end{figure}

The ranking of guidance models by scenario, illustrated in \cref{fig:pgm_classification}, reveals that no modality is systematically the most preferred, although the guiding tube model is generally well-ranked, particularly in scenarios $\bm{S2}$ and $\bm{S4}$. In contrast, the potential field modality is generally the least preferred by participants, except in scenario $\bm{S1}$ where it is the most appreciated. This overall negative perception could be due to the force discontinuity at task initialisation. \added{While this discontinuity is inherent to the model}, gradually activating the guidance force at the beginning of each task, \added{or starting far from any attractive or repulsive points,} could mitigate this issue in future implementations.\added{Although this might lower the global ranking of the potential field model, it remains the most preferred model in the free-environment scenario $\bm{S1}$, suggesting that this does not affect the model's positive perception.}

\begin{figure}[!ht]
    \centering
    \begin{subfigure}[t]{0.49\textwidth}
        \centering
        \includegraphics[width=0.8\textwidth]{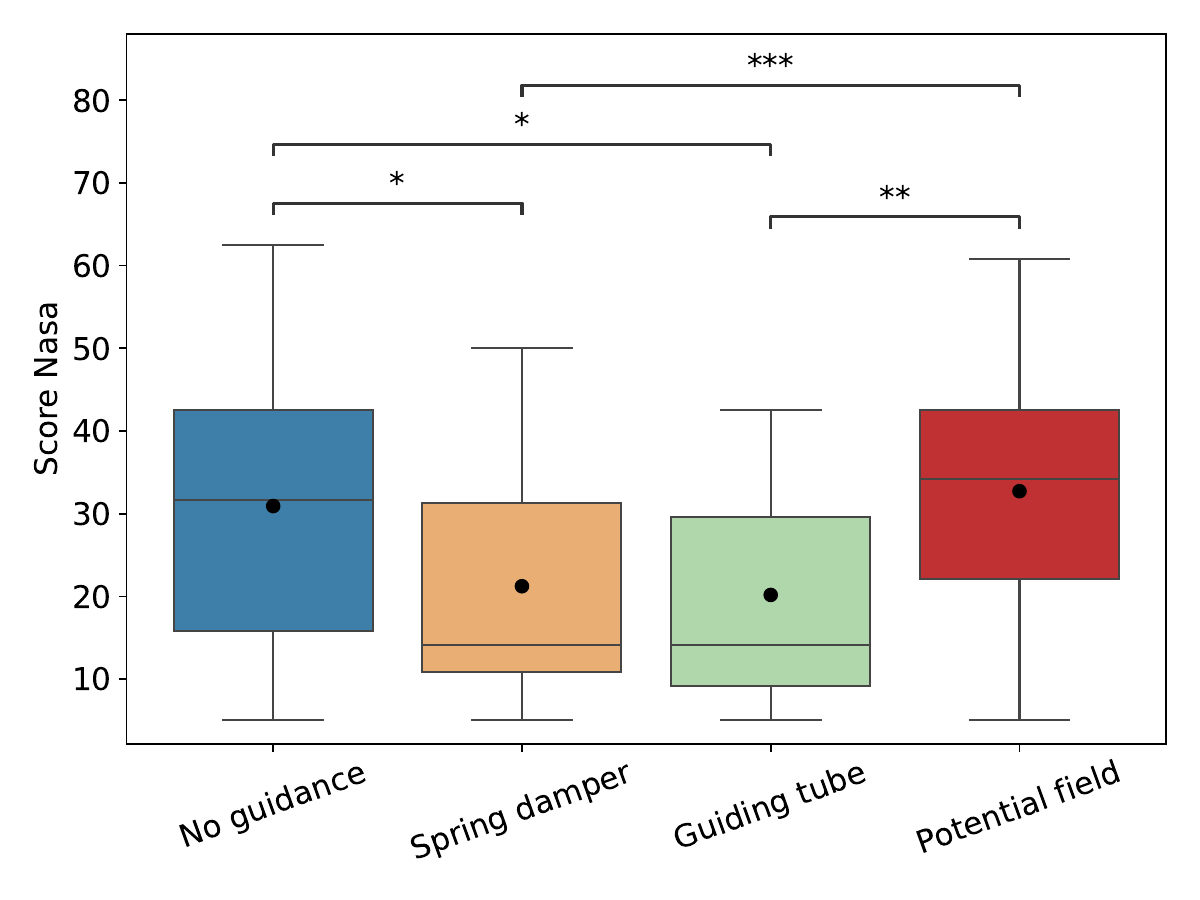}
        \caption{\added{NASA-RTLX scores for all guidance models in scenario $\bm{S4}$}}
        \label{fig:nasa_s4}
    \end{subfigure}
    \hfill
    \begin{subfigure}[t]{0.49\textwidth}
        \centering
        \includegraphics[width=\textwidth]{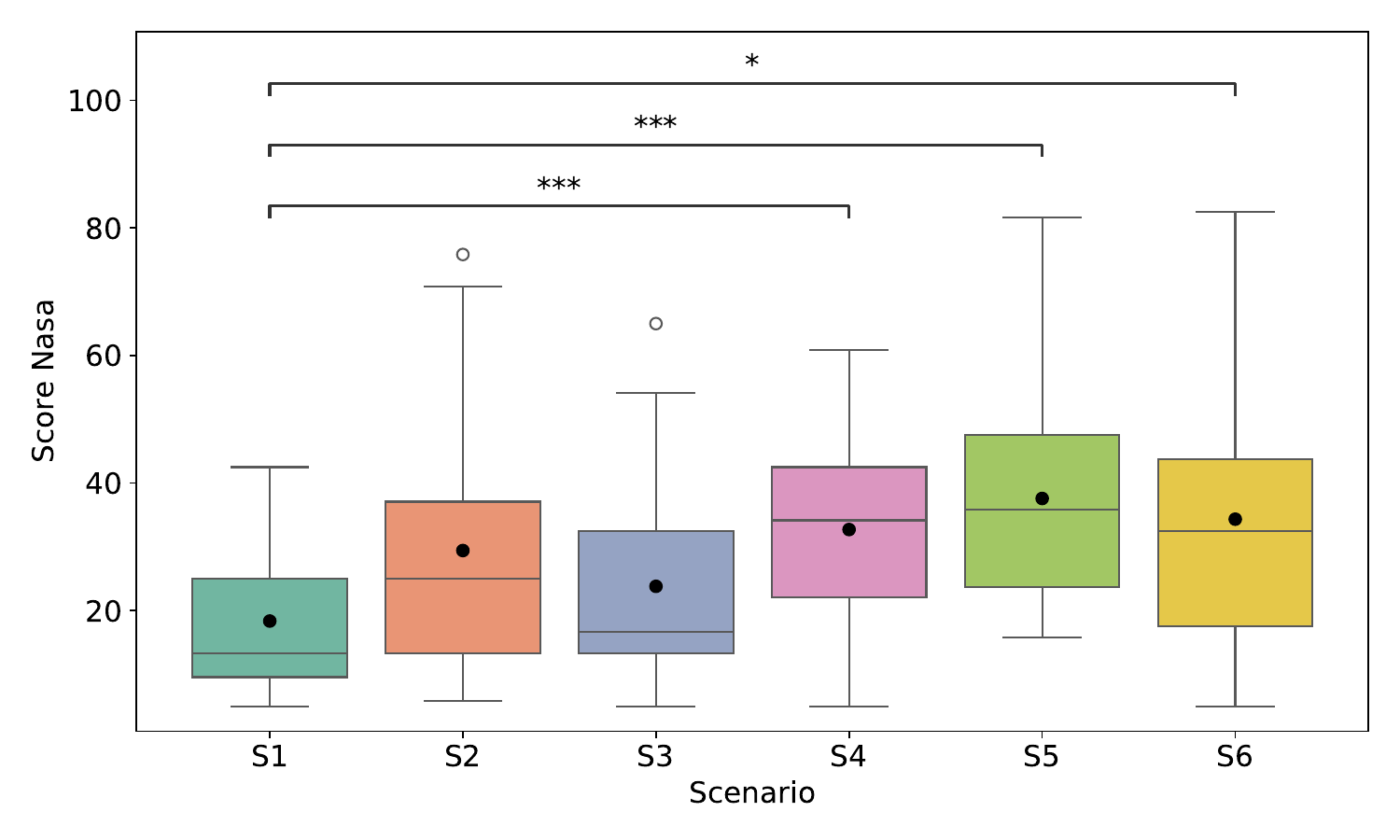}
        \caption{\added{NASA-RTLX scores for potential field model across all scenarios}}
        \label{fig:nasa_pf}
    \end{subfigure}
    \caption{\added{Significant differences are indicated by asterisks (* $p<0.05$, ** $p<0.01$, *** $p<0.001$ and **** $p<0.0001$) and are based on the mean values (black dots)}}\label{fig:nasa_scores}
\end{figure}

\added{Regarding NASA-RTLX global scores, significant differences between guidance models were found, mainly in scenario $\bm{S4}$, which is the most demanding to complete. In this scenario, as illustrated in \cref{fig:nasa_s4}, the spring-damper and guiding tube models reduce mental workload compared to the potential field model ($p = 4.3 \times 10^{-4}$ and $p = 2.0 \times 10^{-3}$ resp.) and the no guidance modality ($p = 0.045$ and $p = 0.023$ resp.) which is consistent with the ranking results.
Focusing on the potential field model across all scenarios, illustrated in \cref{fig:nasa_pf}, we observe a significantly lower mental workload in scenario $\bm{S1}$ compared to scenario $\bm{S4}$, $\bm{S5}$ and $\bm{S6}$ ($p = 1.0 \times 10^{-3}$, $p = 2.3 \times 10^{-4}$ and $p = 0.018$ resp.).}

\begin{figure}[!ht]
    \centering
    \includegraphics[width=0.8\textwidth]{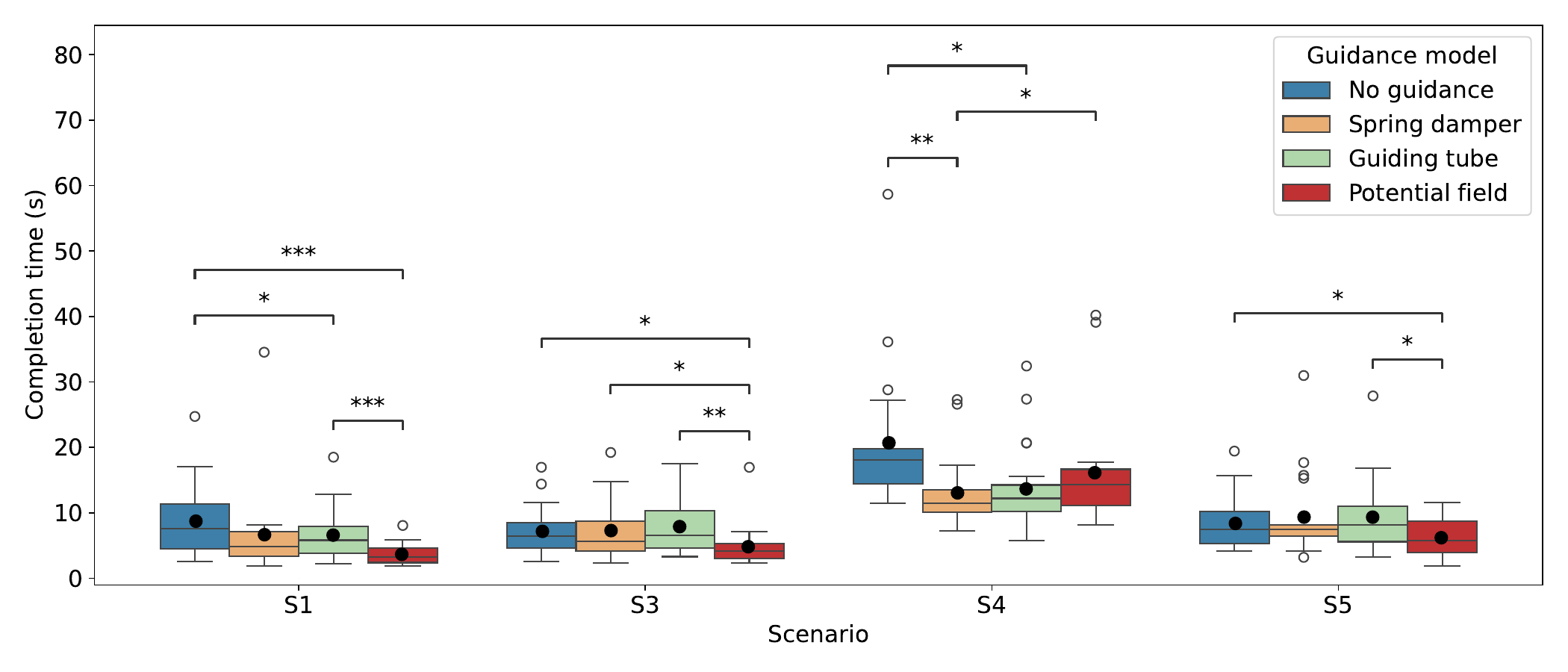}
    \caption{\added{Completion time across guidance models in all scenarios except $\bm{S2}$ and $\bm{S6}$ where no significant differences were found. Significant differences are based on the mean values (black dots)}}\label{fig:completion_time}
\end{figure}

\added{Regarding completion time ($\bm{M_{P_1}}$) of each scenario, as illustrated in \cref{fig:completion_time}, we observe in scenario $\bm{S4}$ a significantly lower time with the spring-damper model compared to no guidance ($p = 4.0 \times 10^{-3}$) and to the potential field model ($p = 0.040$). However, using the potential field model leads to a significantly lower time than both no guidance and the guiding tube model in scenario $\bm{S1}$ ($p = 5.2 \times 10^{-4}$ and $p = 1.0 \times 10^{-3}$ resp.) and scenario $\bm{S5}$ ($p = 0.039$ and $p = 0.043$ resp.). In scenario $\bm{S3}$ completion time is significantly lower for the potential field model than for no guidance, the guiding tube, and the spring damper models ($p = 0.028$, $p = 7.0 \times 10^{-3}$ and $p = 0.012$ resp.).}

\begin{figure}[!ht]
    \centering
    \includegraphics[width=0.8\textwidth]{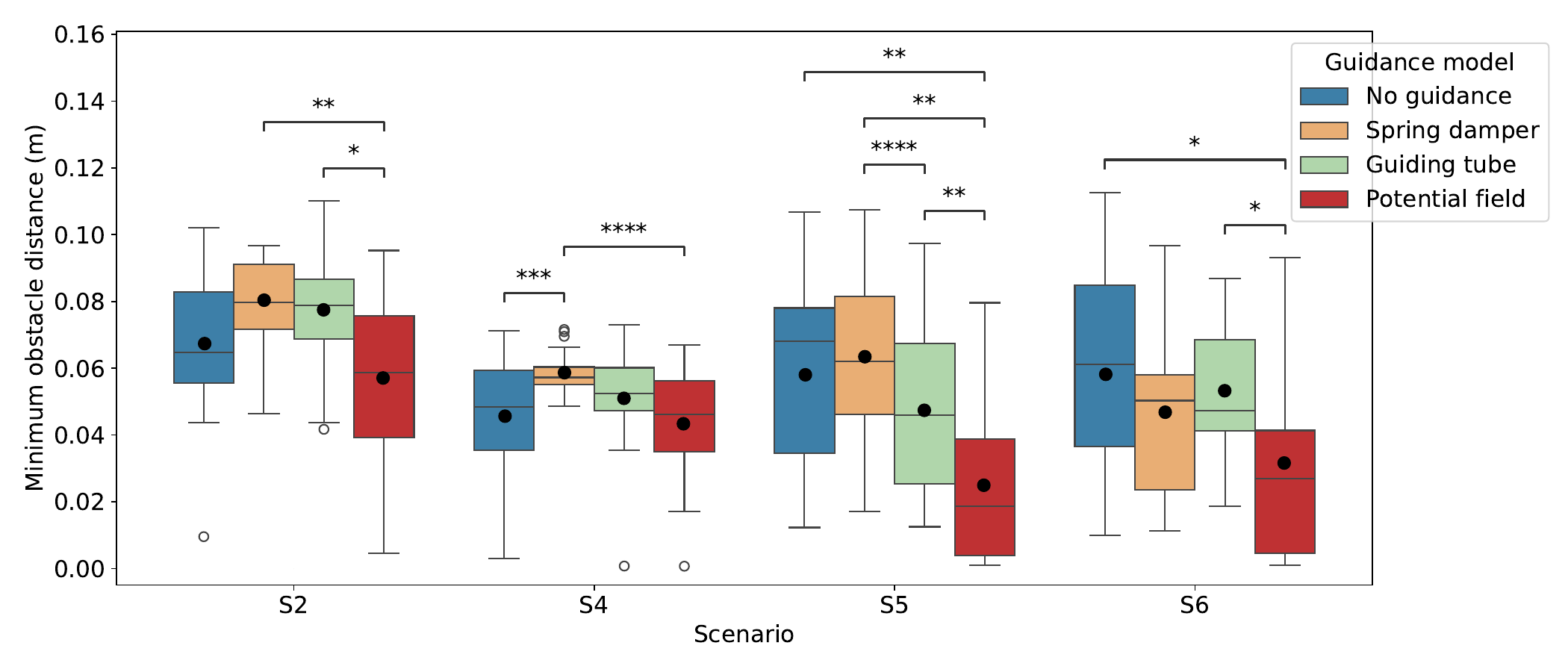}
    \caption{\added{Minimum distance to obstacles across guidance models in all scenarios except $\bm{S1}$ where there is no obstacle to consider and $\bm{S3}$ where no significant differences were found. Significant differences are based on the mean values (black dots)}}\label{fig:obstacle_distance}
\end{figure}

\added{Regarding minimum distance to obstacles ($\bm{M_{S}}$) of each scenario, as illustrated in \cref{fig:obstacle_distance}, we observe that the potential field model leads to significantly smaller distances than the spring-damper model in scenarios $\bm{S2}$ ($p = 6.0 \times 10^{-3}$), $\bm{S4}$ ($p = 4.0 \times 10^{-3}$), and $\bm{S5}$ ($p = 3.3 \times 10^{-5}$), and than the guiding tube model in scenarios $\bm{S2}$ ($p = 0.019$), $\bm{S5}$ ($p = 0.012$), and $\bm{S6}$ ($p = 0.022$).}

These results show that the spring-damper model suits cluttered environments ($\bm{S4}$), as it leads to lower cognitive load, faster completion time, and safer (larger distance to obstacle) execution of the task in comparison to the potential field model and unguided teleoperation. The potential field model suits free spaces ($\bm{S1}$) as it leads to faster completion time than the guiding tube model and the unguided teleoperation. It leads to lower cognitive load in those environments than in cluttered environments ($\bm{S4}$) but increases the risk of collision in environments with obstacles. The guiding tube model offers a balanced compromise between cognitive comfort and safety but leads to slower task completion than the potential field method. In poorly defined environments when the tube radius is insufficient to avoid the obstacle ($\bm{S5}$), performance degrades. \added{Recommendations for guidance model selection based on environmental characteristics are summarized in \cref{tab:model_selection_summary}.}

\begin{table}[!ht]
\centering
% \color{red}
% \arrayrulecolor{red}
\caption{\added{Environment-to-model summary from study results.}}\label{tab:model_selection_summary}
\footnotesize
\begin{tabular}{|>{\centering\arraybackslash}p{0.52\textwidth}|>{\centering\arraybackslash}p{0.42\textwidth}|}
\hline
\textbf{Environmental feature} & \textbf{Recommended model} \\
\hline
Free/open space ($\bm{S1}$) & Potential field \\
\hline
Moderate clutter ($\bm{S2}$) & Guiding tube \\
\hline
Highly cluttered space ($\bm{S4}$) & Spring-damper \\
\hline
\end{tabular}
\end{table}

Both subjective (NASA-RTLX, guidance ranking) and objective (completion time, distance to obstacles) metrics indicate that no model is overall better than the others, and that the choice of the guidance model must be made considering the task and environment characteristics.

\added{Although our evaluation is based on a vertical farming task, this environment-dependent selection logic is consistent with other teleoperation domains commonly studied in haptics. In medical teleoperation (e.g., telesurgery or tele-diagnosis), the spring-damper model may be better suited to constrained maneuvers near sensitive anatomical structures, while potential-field assistance may remain useful during less constrained approach phases. In industrial teleoperation (e.g., remote inspection or maintenance), guiding tube constraints can similarly provide a compromise between operator freedom and safety margins in cluttered workspaces. The proposed model-selection guidelines and interaction metrics can be applied beyond agriculture; the main difference lies in the choice of gains, which depends on task requirements and user preferences.}

\subsection{Evaluation of human-guidance interaction metrics}

\removed{Scenarios $\bm{S5}$ and $\bm{S6}$ were designed to study the limits of the different models under poorly defined guidance. Compared to other scenarios, they lead to significantly lower subjective scores regardless of the guidance model used: for the NASA-RTLX performance question ($\bm{S5}<\bm{S1}, \bm{S2},\bm{S4}$ with $p = 1.0 \times 10^{-3}$, $0.034$, $0.023$ and $\bm{S6}<\bm{S1}$ with $p = 0.017$), for the NASA-RTLX comfort question ($\bm{S5}, \bm{S6}<\bm{S1}$ with $p = 4.0 \times 10^{-3}, 1.6 \times 10^{-4}$), and for trust in the Muir questionnaire ($\bm{S5}<\bm{S1}, \bm{S2}, \bm{S4}$ with $p = 5.05 \times 10^{-8}, 5.21 \times 10^{-5}, 2.66 \times 10^{-4}$ and $\bm{S6}<\bm{S1}, \bm{S2}, \bm{S4}$ with $p = 8.09 \times 10^{-9}, 5.0 \times 10^{-3}, 5.90 \times 10^{-4}$).}

\added{Scenarios $\bm{S5}$ and $\bm{S6}$ were designed to study the limits of the different models under poorly defined guidance. Compared to other scenarios, they lead to significantly lower subjective scores regardless of the guidance model used as summarized in \cref{tab:s5s6_subjective_pvalues}.}

\begin{table}[!ht]
\centering
% \color{red}
% \arrayrulecolor{red}
\caption{\added{Subjective-score differences in poorly defined scenarios ($\bm{S5}$, $\bm{S6}$): significant comparisons and associated $p$-values.}}\label{tab:s5s6_subjective_pvalues}
\footnotesize
\begin{tabular}{|>{\centering\arraybackslash}p{0.22\textwidth}|>{\centering\arraybackslash}p{0.36\textwidth}|>{\centering\arraybackslash}p{0.36\textwidth}|}
\hline
\textbf{Measure} & \textbf{Comparison with $\bm{S5}$ ($p$-value)} & \textbf{Comparison with $\bm{S6}$ ($p$-value)} \\
\hline
NASA-RTLX performance & $\bm{S5}<\bm{S1}$ ($1.0 \times 10^{-3}$)\newline $\bm{S5}<\bm{S2}$ ($0.034$)\newline $\bm{S5}<\bm{S4}$ ($0.023$) & $\bm{S6}<\bm{S1}$ ($0.017$) \\
\hline
NASA-RTLX comfort & $\bm{S5}<\bm{S1}$ ($4.0 \times 10^{-3}$) & $\bm{S6}<\bm{S1}$ ($1.6 \times 10^{-4}$) \\
\hline
Muir trust & $\bm{S5}<\bm{S1}$ ($5.05 \times 10^{-8}$)\newline $\bm{S5}<\bm{S2}$ ($5.21 \times 10^{-5}$)\newline $\bm{S5}<\bm{S4}$ ($2.66 \times 10^{-4}$) & $\bm{S6}<\bm{S1}$ ($8.09 \times 10^{-9}$)\newline $\bm{S6}<\bm{S2}$ ($5.0 \times 10^{-3}$)\newline $\bm{S6}<\bm{S4}$ ($5.90 \times 10^{-4}$) \\
\hline
\end{tabular}
\end{table}

Examining our objective metrics, we find that the mean path-tracking accuracy (performance metric $\bm{M_{P_2}}$ for spring-damper and guiding tube models) and the mean force-tracking accuracy (trust metric $\bm{M_{T_2}}$ for the potential field model) do not capture this degradation in scenarios $\bm{S5}$ and $\bm{S6}$. This invalidates their use as metrics to assess interaction quality.

\begin{figure}
    \centering
    \includegraphics[width=0.8\textwidth]{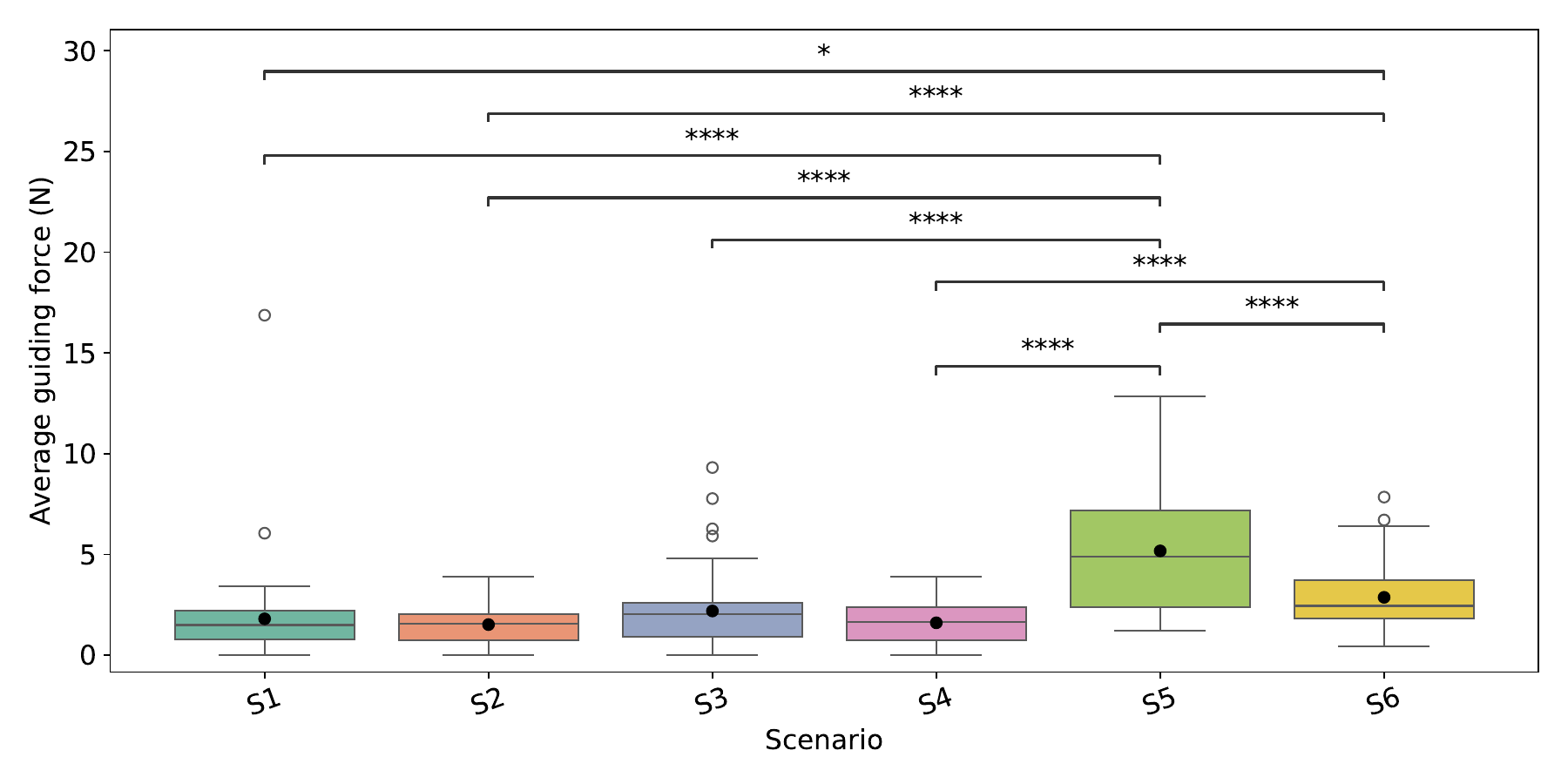}
    \caption{\added{Mean guiding force magnitude ($M_C$ and $M_{T_1}$) across all scenarios. Significant differences are indicated by asterisks (*) and are based on the mean values (black dots)}}\label{fig:average_guiding_forces}
\end{figure}

However, we observe a significant increase in the mean guiding force magnitude (comfort metric $\bm{M_{C}}$ for all guidance models and trust metric $\bm{M_{T_1}}$ for spring-damper and guiding tube models) in these scenarios compared to others, regardless of the guidance model, \added{as depicted in \cref{fig:average_guiding_forces}.}

The mean guiding force magnitude is significantly higher in scenario $\bm{S5}$ than in scenarios $\bm{S1}$ ($p = 7.06 \times 10^{-8}$), $\bm{S2}$ ($p = 1.5 \times 10^{-9}$), $\bm{S3}$ ($p = 8.7 \times 10^{-7}$), and $\bm{S4}$ ($p = 1.05 \times 10^{-8}$). It is also higher in scenario $\bm{S6}$ than in scenarios $\bm{S1}$ ($p = 0.048$), $\bm{S2}$ ($p = 9.44 \times 10^{-7}$), and $\bm{S4}$ ($p = 2.55 \times 10^{-5}$).

This reveals a strong correlation between high guiding forces and low perceived comfort/trust scores from standard questionnaires (NASA-RTLX and Muir), suggesting that guiding force magnitude, and particularly comfort evaluation, could be an effective metric for assessing human-guidance interaction quality.
While the mean values of path and force-tracking accuracies appear less relevant as post-task metrics, they may prove valuable when evaluated in real-time during teleoperation and deserve further investigation. \added{Using guiding force magnitude as a real-time indicator to adapt the guidance model and optimize user comfort is also promising.}

\section{Conclusion}
\label{sec:conclusion}

% \removed{We propose an overview of the main haptic guidance models that can be used in teleoperation tasks. We define a unified formalism to express those haptic guidance models as a stiffness-damping system and compare them. We conducted a user study which highlights that no haptic guidance model is universally superior; each has strengths and weaknesses in terms of performance, safety, comfort and trust, depending on the task and environment. Several guidelines for selecting appropriate guidance models are derived from the results: the spring-damper model excels in cluttered settings, while the potential field model is better suited for free environments. The guiding tube model offers a balanced compromise but may be less efficient in certain scenarios. To evaluate human-guidance interaction, we find out that comfort-related metrics, particularly guiding force magnitude, effectively reflect user experience. 
% Future work will focus on adaptive haptic guidance, which could dynamically switch between guidance models to optimize user comfort using generic haptic guidance models~\cite{boulay2024_ruling} and the real-time evaluation metrics.
% }
We propose an overview of the main haptic guidance models that can be used in teleoperation tasks. We define a unified \added{formulation} to express these haptic guidance models as a stiffness-damping system and compare them. We conducted a \added{comparative} user study showing that no haptic guidance model is universally superior; each has strengths and weaknesses in terms of performance, safety, comfort, and trust, depending on the task and environment. \added{Based on these results, we propose} guidelines for selecting appropriate guidance models: the spring-damper model excels in cluttered settings, while the potential field model is better suited for free environments. The guiding tube model offers a balanced compromise but may be less efficient in certain scenarios. To evaluate human-guidance interaction, we test several metrics and find that comfort-related metrics, particularly guiding force magnitude, effectively reflect user experience.
Future work will focus on adaptive haptic guidance models, allowing dynamic switching between guidance models \added{and on the use of adaptive virtual objects}, to optimize user comfort using real-time evaluation metrics \added{derived from those proposed in this paper}.

-----------------------------------------------
-----------------------------------------------

\begin{credits}
\subsubsection{\ackname} This work is supported by an ANRT CIFRE fellowship granted to Farm3. The authors thank Lucas Joseph and Benjamin Camblor for their help in designing the experiment.
% \subsection{\ackname} Funding information and other acknowledgments are anonymized for initial submission.

\subsubsection{\discintname}
The authors have no competing interests to declare that are relevant to the content of this article.
\end{credits}
\bibliographystyle{splncs04}
\bibliography{references}
\end{document}